\documentclass{article}

\usepackage{microtype}
\usepackage{graphicx}
\usepackage{subfigure}
\usepackage{booktabs} 

\usepackage{hyperref}

\usepackage[accepted]{icml2019}

\usepackage{amsmath}
\usepackage{bbold}
\numberwithin{equation}{section}
\numberwithin{figure}{section}
\usepackage{authblk}
\usepackage{hyperref}

\newcommand*{\Bm}[1]{\boldsymbol{#1}}
\newcommand*{\Bv}[1]{\boldsymbol{#1}}

\newcommand{\unitary}{orthogonal }

\newtheorem{theorem}{Theorem}

\icmltitlerunning{Rotation Invariant Householder Parameterization for Bayesian PCA}

\begin{document}

\twocolumn[
\icmltitle{Rotation Invariant Householder Parameterization for Bayesian PCA}




\begin{icmlauthorlist}
\icmlauthor{Rajbir S. Nirwan}{to}
\icmlauthor{Nils Bertschinger}{to,goo}
\end{icmlauthorlist}

\icmlaffiliation{to}{Department of Computer Science, Goethe University, Frankfurt, Germany}
\icmlaffiliation{goo}{Frankfurt Institute for Advanced Studies, Frankfurt, Germany}

\icmlcorrespondingauthor{Rajbir S. Nirwan}{nirwan@fias.uni-frankfurt.de}
\icmlcorrespondingauthor{Nils Bertschinger}{bertschinger@fias.uni-frankfurt.de}

\icmlkeywords{Bayesian Machine Learning, PCA}

\vskip 0.3in
]



\printAffiliationsAndNotice{}

\begin{abstract}
  We consider probabilistic PCA and related factor models from a
  Bayesian perspective. These models are in general not identifiable
  as the likelihood has a rotational symmetry. This gives rise to
  complicated posterior distributions with continuous subspaces of
  equal density and thus hinders efficiency of inference as well as
  interpretation of obtained parameters. In particular, posterior
  averages over factor loadings become meaningless and only model
  predictions are unambiguous. Here, we propose a parameterization
  based on Householder transformations, which remove the rotational
  symmetry of the posterior. Furthermore, by relying on results from
  random matrix theory, we establish the parameter distribution which
  leaves the model unchanged compared to the original rotationally
  symmetric formulation. In particular, we avoid the need to compute
  the Jacobian determinant of the parameter transformation. This
  allows us to efficiently implement probabilistic PCA in a rotation
  invariant fashion in any state of the art toolbox. Here, we
  implemented our model in the probabilistic programming language Stan
  and illustrate it on several examples.
\end{abstract}

\section{Introduction}

Modern algorithms and computational tools have vastly expanded the scope of Bayesian
modeling over the last decades. In particular, Hamiltonian Monte-Carlo
(HMC) sampling (see \citet{Betancourt2017} for a thorough introduction)
and variational inference \citep{PRML} allow to approximate
high-dimensional posterior distributions. In addition, software tools
such as probabilistic programming languages, e.g. Stan \citep{Stan}, or
libraries for automatic differentiation, e.g. Tensorflow \citep{tensorflow2015-whitepaper},
simplify the implementation. Thanks to these advances, researchers can
focus on the statistical modeling as model implementation often
requires just a few lines of code. Nevertheless, some models remain
challenging. Well known examples include Probabilistic PCA (PPCA) and
related factor models which are widely used in exploratory and
confirmatory data analysis. These models are non-identifiable, which
poses major problems when fitted parameters, e.g. factor loadings, are
being interpreted. In this context, the non-identifiability arises from a
rotational symmetry of the likelihood model. While identification can
be restored by imposing constraints on the factor loadings
\citep{Joereskog1969,Peeters2012}, maximum likelihood estimation can be
biased by the imposed constraints \citep{Millsap2001}.

Here, we consider PPCA from a Bayesian perspective.
In this context, the rotational symmetry gives rise to a complicated
posterior with continuous subspaces of equal density.  Continuous
symmetries can severely reduce sampling efficiency as many samples are 
required in order to explore the corresponding subspaces.
Furthermore, interpretation of the sampled parameters becomes
impossible as marginal posterior averages are meaningless when many
different parameter combinations give rise to the same
likelihood. This is akin to the discrete label switching symmetry
observed in mixture models. In this case, the parameters of the individual
mixture components cannot be averaged as the arbitrary component
labels between different samples could be switched. Thus, a major
advantage of Bayesian modeling, namely the ability to access estimation
uncertainty, cannot be utilized as posterior means and variances not
just reflect variation within, but also between components.

Therefore, it is desirable to remove the rotational symmetry of PPCA.
Formally, the symmetry arises due to the choice of
coordinate system for the principal components being arbitrary. In
particular, a rotation of the coordinate axes leaves the model
likelihood unchanged. Mathematically, the rotation symmetry is made
explicit by formulating the model in terms of an orthogonal matrix
(transforming into a fixed coordinate system) and a diagonal matrix containing the
explained variances. To this end, we employ the singular value
decomposition (SVD) and parameterize the Stiefel manifold of
orthogonal matrices in terms of non-redundant unconstrained
parameters. The mapping between parameters and orthogonal matrices is
provided in terms of a sequence of Householder transformations.

\subsection{Related work}

Special sampling methods for inference on statistical manifolds have
been developed, in which the geometric structure of the manifold
is respected and geodesic trajectories are traced out with respect to
the metric of the manifold \citep{Byrne13}. In general, the geometry
will be non-Euclidean, e.g. spherical for unit vectors. Also the
geometry of the Stiefel manifold of orthogonal matrices can be handled
in this fashion. Unfortunately, as the geometry needs to be build into
the sampling algorithm, this approach requires substantive work. 
This also applies to approaches using analytic results for the matrix von Mises-Fisher
distribution on the Stiefel manifold. Exploiting conditional conjugacy
\citep{Hoff2009} and \citep{Smidl2007} have employed this distribution to implement Gibbs sampling
and variational inference for Bayesian PPCA respectively.
An alternative, which allows to utilize general purpose tool boxes, is to
reparameterize the manifold in terms of unconstrained
parameters\footnote{In general, such a reparameterization cannot be
  achieved globally. Instead, the non-Euclidean geometry gives rise to
  singularities in the parameter transformation, e.g. at the north
  pole when mapping a sphere onto a plane, or a different mapping,
  i.e. chart, needs to be employed at different points.}.
\citet{pourzanjani2017} achieved this via Givens rotations. They
also demonstrate that the model can then be implemented in Stan
without any changes to the underlying sampling routine and
unrestricted by conjugacy requirements.


Ideally, reparameterizing a Bayesian model should not change the joint density defined
by the model. In case of the transformation via Givens rotations, the
parameter density corresponding to a uniform measure on the Stiefel
manifold is not known. Thus, by change of measure the density needs to
be corrected by the Jacobian determinant, which poses
a major computational bottleneck in high-dimensional models. 
Here, we propose a different parameterization in terms of Householder
transformations. In this case, results from random matrix theory allow
us to obtain the induced parameter density corresponding to a
Gaussian prior on the original parameters of PPCA, which include the
rotational symmetry. In section \ref{background} we recall
the necessary mathematical results to do so. Then, we
illustrate the resulting model on several examples (section
\ref{UBPCA} and \ref{GPLVM}) before we conclude in section \ref{conclusion}.

\section{Background}
\label{background}



Here, we quickly review PPCA and explain why the model is rotationally
symmetric, and how the symmetry can be removed via singular value decomposition (SVD). Finally, we discuss some results from
random matrix theory in order to construct a suitable prior on the
transformed parameters.

\subsection{Probabilistic Principal Component Analysis (PPCA)}
\label{ppca}

PPCA \citep{Tipping1999,PRML} relates a $D$-dimensional observation
$\Bv{y}$ to a $Q$-dimensional latent vector $\Bv{x}$ via a linear
mapping $\Bm{W} \in \mathbb{R}^{D \times Q}$
\begin{equation}
 \Bv{y} = \Bm{W} \Bv{x} + \Bv{\mu} + \Bv{\epsilon},
 \label{linear_eq_single_y}
\end{equation}
where $\Bv{\mu}$ allows the model to have a non-zero mean and 
$\Bv{\epsilon}$ states the variance not explained by the first 
two terms as noise.
PPCA assumes a standard Gaussian distribution on the latent space, $\Bv{x} \sim 
\mathcal{N}(\Bv{0}, \Bm{I})$ and a zero mean Gaussian noise
with variance $\sigma^2$ for all dimensions, i.e. $\Bv{\epsilon} \sim
\mathcal{N}(\Bv{0}, \sigma^2 \Bm{I})$. The likelihood for $\Bv{y}$ given 
$\Bm{W}, \Bm{\mu}$ and $\sigma^2$ takes the form 
\begin{equation}
 p(\Bv{y}| \Bm{W}) = \mathcal{N}(\Bv{\mu}, \Bm{W}\Bm{W}^T + \sigma^2 \Bm{I}).
 \label{likelihood_single_y}
\end{equation}
For $N$ observations, denoted as $\Bm{Y} = (\Bv{y}_1, \Bv{y}_2, ..., 
\Bv{y}_N)^T \in \mathbb{R}^{N \times D}$, equation (\ref{linear_eq_single_y})
becomes
\begin{equation}
 \Bm{Y} = \Bm{X}\Bm{W}^T + \Bm{\epsilon},
 \label{linear_eq_y}
\end{equation}
where $\Bm{X} = (\Bv{x}_1, \Bv{x}_2, ..., \Bv{x}_N)^T \in 
\mathbb{R}^{N \times Q}$ and $\Bm{\epsilon} \in \mathbb{R}^{N \times D}.$
The likelihood of $\Bm{Y}$ given $\Bm{W}$ then becomes 
\begin{align}
 p(\Bm{Y}|\Bm{W}) &=  \prod_{n=1}^{N} \mathcal{N}(\Bm{Y}_{n,:} | \Bv{\mu}, 
 \Bm{W}\Bm{W}^T + \sigma^2 \Bm{I}), \nonumber \\
 \ln p(\Bm{Y}|\Bm{W}) &=  \frac{ND}{2} \ln(2\pi) - \frac{N}{2} \ln|\Bm{K}| -
 \frac{1}{2} \text{tr} (\Bm{K}^{-1}\tilde{\Bm{Y}}\tilde{\Bm{Y}}^T),
 \label{likelihood_pca}
\end{align}
where $\Bm{K} = \Bm{W} \Bm{W}^T + \sigma^2 \Bm{I}$ and  $\tilde{\Bm{Y}}$ 
are the centered observations, i.e. $\tilde{\Bm{Y}}_{n,:} = \Bm{Y}_{n,:} - 
\Bm{\mu}$. This expression can be maximized analytically 
and yields the solution \citep{Tipping1999}
\begin{align}
  \Bm{\mu}_{\text{ML}} &= \frac{1}{N} \sum_n \Bm{Y}_{n,:} , \nonumber \\
  \Bm{W}_{\text{ML}} &= \Bm{U} \left( \Bm{\Lambda} - \sigma^2 \Bm{I} \right)^{1/2} \Bm{R},
 \label{ml_solution_pca}
\end{align}
where $\Bm{U} \in \mathbb{R}^{D \times Q}$ is an \unitary matrix 
containing the principal eigenvectors of the empirical covariance matrix 
$\frac{1}{N}\tilde{\Bm{Y}}\tilde{\Bm{Y}}^T$, $\Bm{\Lambda} \in 
\mathbb{R}^{Q \times Q}$ is a diagonal matrix containing the
corresponding eigenvalues $(\lambda_1, ..., \lambda_Q)$ 
of $\frac{1}{N}\tilde{\Bm{Y}}\tilde{\Bm{Y}}^T$
on the diagonal and $\Bm{R} \in \mathbb{R}^{Q \times Q}$ is an arbitrary 
rotation matrix. The maximum likelihood estimation for $\sigma^2$ is 
given by
\begin{equation}
 \sigma^2_{\text{ML}} = \frac{1}{D-Q} \sum_{i = Q+1}^D \lambda_i,
\end{equation}
which is basically the variance of $\Bm{Y}$ not picked up by the model.
By projecting the data points $\Bm{y}$ into the latent space, a representation
of reduced dimensionality $Q < D$ can be obtained. The posterior mean
of the latent vectors is given by \citep{PRML}
\begin{align}
  \label{proj_pca}
  \mathbb{E}[\Bm{x} | \Bm{y}] &= \left( \Bm{W}_{\text{ML}}^T \Bm{W}_{\text{ML}} + \sigma^2 \Bm{I} \right)^{-1}
                                \Bm{W}_{\text{ML}}^T (\Bm{y} - \Bm{\mu}_{\text{ML}}) \, .
\end{align}
Note that the rotation symmetry in the likelihood caused by $\Bm{R}$ in 
(\ref{ml_solution_pca}) makes the likelihood non-unique. Two different
solutions $\Bm{W}$ and $\tilde{\Bm{W}} = \Bm{W}\Bm{R}$ will lead to the 
same likelihood, since $\tilde{\Bm{W}}\tilde{\Bm{W}}^T = 
\Bm{W}\Bm{R}\Bm{R}^T\Bm{W}^T = \Bm{W}\Bm{W}^T$.
For this reason, numerical optimization algorithms often converge to
different results when started from different initial conditions. While
model predictions are unaffected by this non-uniqueness, interpretation
of parameters and the projected latent vectors becomes impossible. Especially
the latter is problematic if PPCA is employed as a pre-processing step
to reduce the dimensionality of the data and further analysis might be
sensitive to the arbitrary choice of latent space rotation.
Figure \ref{W_opt_sampled} (left side) shows the maximum 
likelihood result for 1000 different initial conditions for $\Bm{W}$.
One can clearly see the rotation symmetry. 

PPCA is closely related to classical PCA. In particular, as shown by
\citet{Tipping1999}, the classical solution is recovered by letting
$\sigma^2 \to 0$. In this case, the dimensionality reduction is achieved
by the linear projection
\begin{equation}
  \label{proj_classic}
  \Bm{x} = \Bm{\Lambda}^{- \frac{1}{2}} \Bm{U}^T (\Bm{y} - \Bm{\mu}_{\text{ML}})
\end{equation}
where $\Bm{\mu}_{\text{ML}}, \Bm{\Lambda}$ and $\Bm{U}$ are defined as in
(\ref{ml_solution_pca}). The eigenvalues $\lambda_q$ are interpreted
as the variance explained by the $q$-th principal component $\Bm{U}_{:,q}$
in this context.  From (\ref{proj_pca}) the classical solution is
obtained by assuming vanishing noise variance $\sigma^2$ and fixing
the latent space rotation of the maximum likelihood solution at
$\Bm{R} = \Bm{I}$.

Factor analysis is closely related to PPCA as well. In this case, PPCA is
slightly generalized by assuming that the noise is distributed as
$\Bv{\epsilon} \sim \mathcal{N}(\Bv{0}, \Bm{\Psi})$ with diagonal
covariance matrix $\Bm{\Psi}$. In particular, the likelihood has the
same rotational symmetry as PPCA. Here, we just note that all our
derivations in the following sections apply equally to this
model.

\subsection{Bayesian Approach to PPCA}

The fully Bayesian way to solve (\ref{likelihood_pca}) would be to 
impose prior distributions $p(\Bm{\mu}, \Bm{W}, \sigma^2)$ and solve for the posterior
\begin{equation}
 p(\Bm{\mu},\Bm{W},\sigma^2 | \Bm{Y}) = \frac{p(\Bm{Y} | \Bm{\mu},\Bm{W},\sigma^2)p(\Bm{\mu},\Bm{W},\sigma^2)}{p(\Bm{Y})},
\end{equation}
which is not tractable anymore. Therefore, we have to resort to other techniques, e.g.
sampling the posterior.
For a rotation invariant prior $p(\Bm{W})$ (e.g. an isotropic Gaussian) the 
posterior will have the same rotation symmetry as well. Thus, the posterior
has a complicated shape with a continuous subspace of equal density
posing a considerable challenge for sampling algorithms.

\begin{figure}
 \centering
 \includegraphics[width=0.49\textwidth]{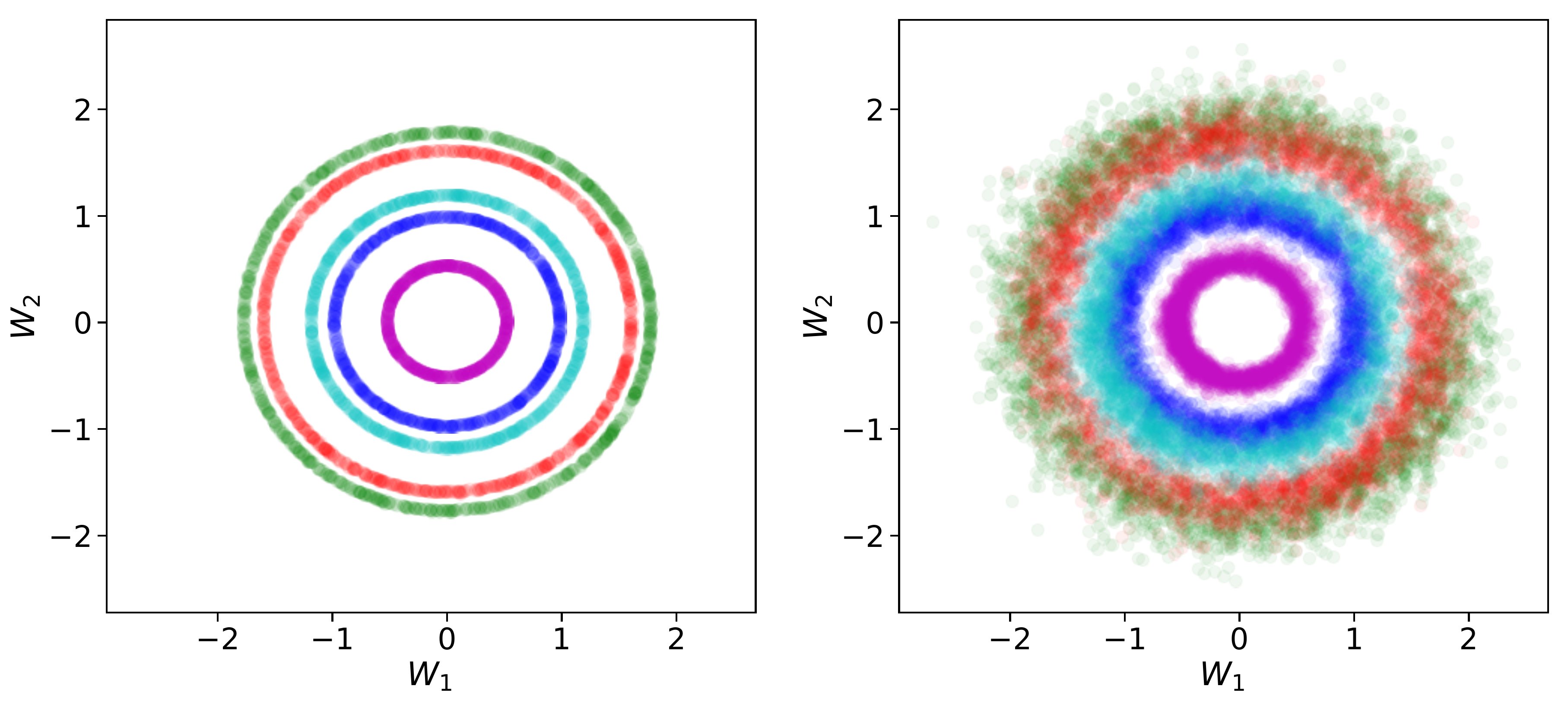}
 \vspace{-0.2cm}
 \caption{Data are created by sampling 50 5-dimensional points from the
 standard normal distribution and map them via a random linear mapping $\Bm{W} \in 
 \mathbb{R}^{5 \times 2}$ to the data-space $\Bm{Y}$. Elements of $\Bm{W}$ are
 drawn from a standard normal distribution. 
 Left side: Maximum likelihood solution for 1000 runs with different 
 initial conditions of $\Bm{W}$. Right side: 4000 samples from the
 posterior distribution. Different colors show different rows of $\Bm{W}$. }
 \label{W_opt_sampled}
\end{figure}

Figure \ref{W_opt_sampled} (right side) shows 4000 posterior samples.
The likelihood is given by (\ref{likelihood_pca}) and the prior on 
$\Bm{W}$ is a standard normal. We can clearly see the rotation symmetry 
again. Samples were drawn using the NUTS sampler of Stan \citep{Stan},
which is able to fully explore the challenging posterior. Yet, in higher
dimensions $D > Q \gg 1$ the problem becomes more severe and a lot of samples
are required to fully explore the symmetric solutions. Furthermore,
statistics of the posterior samples, e.g. computing the mean or variance,
do not have any significance in this case.

The question arises: Can we somehow take out the rotation symmetry without
changing the model (i.e. likelihood or the mass of the posterior distribution)?
The answer is yes:
We can achieve this by reparametrizing $\Bm{W}$. That way
the results are identified uniquely. 
The aim is not to change the model by the reparameterization. Therefore,
we need specific corresponding distributions for the parameters on the
reparameterized model. In the next sections we explain how to do that.

\subsection{Singular Value Decomposition}


It is well known, that every positive-definite matrix $\Bm{S}$ can be diagonalized
by an \unitary transformation $\Bm{U}$ as $\Bm{S} = \Bm{U}\Bm{\Lambda}\Bm{U}^T$,
where $\Bm{\Lambda}$ contains the eigenvalues of $\Bm{S}$ on the diagonal and 
the columns of $\Bm{U}$ are the corresponding normalized eigenvectors.

This is not true anymore for other matrices, but there exists a more
general decomposition called the singular value decomposition (SVD).
A matrix $\Bm{A} \in \mathbb{R}^{D \times Q}$ can be decomposed as
\begin{equation}
 \Bm{A} = \Bm{U} \Bm{\Sigma} \Bm{V}^T,
\end{equation}
where $\Bm{U} = (\Bv{u}_1, ... , \Bv{u}_D)^T, \Bm{V} =
(\Bv{v}_1, ... , \Bv{v}_Q)$ are \unitary matrices and $\Bm{\Sigma} = 
\text{diag}(\sigma_1, ... , \sigma_Q) \in 
\mathbb{R}^{D \times Q}$ is a diagonal matrix containing the singular values.
When $\Bm{A}$ has rank $P < Q$, only $P$ of them will be non-zero.
It is easy to show that the square of the diagonal elements 
$(\sigma_1^2, ... ,\sigma_P^2)$ are the non-zero eigenvalues of $\Bm{A}\Bm{A}^T$
and $\Bm{A}^T\Bm{A}$, which are similar matrices,
and all $\Bv{u}_j$'s and $\Bv{v}_j$'s obey the relation
\begin{align}
 \Bm{A}\Bv{v}_j &= \sigma_j\Bm{u}_j,\ \ \Bm{A}^T\Bv{u}_j = 
 \sigma_j\Bm{v}_j,\ \ \forall \ j = 1, ... , P \nonumber \\
 \Bm{A}\Bv{v}_j &= \Bv{0},\qquad \Bm{A}^T\Bv{u}_j = 
 \Bv{0},\qquad \forall \ j > P
\end{align}

We will decompose our mapping $\Bm{W} \in \mathbb{R}^{D \times Q}$ via SVD as
$\Bm{W} = \Bm{U} \Bm{\Sigma} \Bm{V}^T$. The likelihood
only depends on the outer product of $\Bm{W}$.
Since the outer product does not depend on $\Bm{V}$,
\begin{align}
 \Bm{W}\Bm{W}^T 
 = \Bm{U} \Bm{\Sigma} \Bm{V}^T (\Bm{U} \Bm{\Sigma} \Bm{V}^T)^T 
 = \Bm{U} \Bm{\Sigma}^2 \Bm{U}^T,
 \label{wishartDecomposition}
\end{align}

we can neglect $\Bm{V}$
\footnote{Note, $\Bm{V}$ is precisely 
the rotation symmetry that we want to get rid of.},
i.e., we assume it to be the identity matrix.
That fixes the
coordinate frame of the latent space and takes out the rotation symmetry.

By reparameterizing $\Bm{W}$ by $\Bm{U}$ and $\Bm{\Sigma}$, we
obtain a unique representation\footnote{In general, the SVD is not
  unique. While the singular values can be ordered to ensure
  uniqueness of $\Bm{\Sigma}$, the left singular vectors $\Bm{U}$ are
  only determined up to the sign. As explained later, we enforce a sign
  convention on them.}. 
However, in the fully Bayesian case, we have
a prior on $\Bm{W}$.  After reparameterization by $\Bm{U}$ and
$\Bm{\Sigma}$, we have to make a Jacobian adjustment to the
probability density, since otherwise the prior density mass on
$\Bm{W}$ will change as well.  In the next sections, we consider a
zero mean Gaussian prior on $\Bm{W}$, and how to adjust the prior on
$\Bm{U}$ and $\Bm{\Sigma}$, such that we do not change the distribution on
$\Bm{W}$.

Similarly, SVD provides a numerically stable way for classical PCA.
In this case, the data matrix $\Bm{Y}$ is decomposed as
$\Bm{Y} = \Bm{U}_{PCA} \Bm{\Sigma}_{PCA} \Bm{V}^T_{PCA}$ and the
desired dimensionality reduction is achieved by the linear map
$\Bm{\Sigma}_{PCA} \Bm{U}^T_{PCA}$. Again, we see that the matrix of
right singular vectors $\Bm{V}_{PCA}$, corresponding to an arbitrary choice
of latent coordinate rotation, is dropped.

\subsection{Random Matrix Theory, Haar Measure}
\label{first_part}

SVD decomposes our $\Bm{W}$ into an \unitary matrix $\Bm{U}$, a
diagonal matrix $\Bm{\Sigma}$ and another \unitary matrix $\Bm{V}$, 
that we are neglecting in our analysis, since the outer 
product $\Bm{WW}^T$ does not depend on $\Bm{V}$. 

For a standard Gaussian prior on $\Bm{W}$,
i.e. all elements of $\Bm{W}$ are zero mean Gaussian with unit variance,
$\Bm{WW}^T$ has a Wishart distribution
and the following theorem holds
\citep{James2014}:

\begin{theorem}
 Let the entries of $\Bm{W} \in \mathbb{R}^{D \times Q}$ be i.i.d. Gaussian
 with zero mean and unit variance. The joint probability density of the ordered strictly
 positive eigenvalues of the Wishart matrix $\Bm{W}^T\Bm{W}$, $\lambda_1 \geq ... 
 \geq \lambda_Q$, equals
 \begin{equation}
  p(\Bv{\lambda}) = c e^{- \frac{1}{2} \sum_{q=1}^Q \lambda_q}
  \prod_{q=1}^Q \left( \lambda_q^{\frac{D-Q-1}{2}}
  \prod_{q'=q+1}^Q | \lambda_{q} - \lambda_{q'} | \right),
 \end{equation}
 where $\Bv{\lambda}=(\lambda_1, ...,  \lambda_Q)$ and
 $c$ is a constant that depends on $D$ and $Q$.  
\end{theorem}

Note that the non-zero eigenvalues of $\Bm{W}^T\Bm{W}$
are the same as for $\Bm{W}\Bm{W}^T$, and therefore they have the
same probability density function.
The singular values $\sigma_i$ of 
$\Bm{W}$ are the square roots of $\Bv{\lambda}$'s, 
$\sigma_i = \sqrt{\lambda_i}$.
Thus, by change of measure we get\footnote{Given a probability density $p_x(x)$ on x and
an invertible map $g$, so that $y = g(x)$, the density
$p_y(y)$ on $y$ is given by $p_y(y) = p_x(g^{-1}(y)) 
\left|\text{det}\left(\frac{dg^{-1}(y)}{dy}\right)\right|.$
Here the absolute determinant of the Jacobian accounts for the change
of volume under $g$.}
\begin{align}
  \label{pdf_sigma}
  p&(\sigma_1, ..., \sigma_Q) = \nonumber \\ 
  &c e^{- \frac{1}{2} \sum_{q=1}^Q \sigma_q^2}
    \prod_{q=1}^Q \left( \sigma_q^{D-Q-1} 
    \prod_{q'=q+1}^Q | \sigma_q^2 - \sigma_{q'}^2 | \right)
    \prod_{q=1}^Q 2 \sigma_q,
\end{align}
where the last product in the above term is the Jacobian correction.
Provided this density function as the prior distribution on the
singular value matrix $\Bm{\Sigma}$, we can easily sample from its
posterior.

To calculate the prior distribution for $\Bm{U}$, we need
to dig deeper into random matrix theory.
As mentioned earlier, a standard Gaussian prior on $\Bm{W}$
gives rise to a Wishart distribution on $\Bm{W}\Bm{W}^T$.
Equation (\ref{wishartDecomposition}) shows, that a 
Wishart matrix can be decomposed into a product of 
an \unitary matrix $\Bm{U}$ and a diagonal matrix $\Bm{\Sigma}$.
Furthermore, it is known that the eigenvectors
$\Bm{U}$ are distributed uniformly in the space of 
\unitary matrices\footnote{\citet{bai2007} and 
\citet{uhlig1994} mention that the 
eigenvector matrix
of a Wishart distribution is Haar-distributed, which is 
also true for singular Wishart matrices, but we 
could not find a proof anywhere. 
}.
The set of \unitary matrices is known as the Stiefel manifold
$\mathcal{V}_{Q, D}$
\begin{equation}
  \label{def_Stiefel}
  \mathcal{V}_{Q, D} = \left\{ \Bm{U} \in \mathbb{R}^{D \times Q} | \Bm{U} \Bm{U}^T = \Bm{I} \right\} \, .
\end{equation}
The dimension of this manifold is $D Q - \frac{1}{2} Q (Q + 1)$,
accounting for the fact that the orthogonality constraint reduces the
number of independent degrees of freedom. The Stiefel manifold can be
equipped with a uniform measure. This measure is an example of a {\em
  Haar} measure, as it is invariant under the action of the orthogonal
group $O(D) \simeq \mathcal{V}_{D,D}$, i.e.
\begin{equation}
  \label{Haar}
  p(\Bm{U}) = p(\Bm{R} \Bm{U}) \quad \forall \Bm{R} \in O(D) \, .
\end{equation}

To proceed, we need to find an unconstrained parameterization for \unitary matrices
along with a density on the parameters, such that the resulting 
matrix is Haar distributed. \citet{ron2014, pourzanjani2017} and \citet{mezzadri2007} 
suggest ways to do that. 
The procedure is explained in detail in the next section.

\subsection{QR-decomposition and Householder Transformations}
\label{second_part}
Given a matrix $\Bm{Z} \in \mathbb{R}^{D \times Q}$ the
(thin) QR-decomposition decomposes $\Bm{Z}$ into an \unitary matrix
$\Bm{Q} \in \mathcal{V}_{Q,D}$ and an upper triangular matrix $\Bm{R} \in \mathbb{R}^{Q \times Q}$.
If the elements of $\Bm{Z}$ are i.i.d Gaussian with mean
zero and unit variance, $\Bm{Q}$ is Haar distributed \citep{mezzadri2007}. 
Note that this is only the case if a unique QR-decomposition 
is used. In practice, this is usually achieved by enforcing the
convention that the diagonal elements of $\Bm{R}$ are
positive.

To compute the QR-decomposition of $\Bm{Z}$, the so-called Householder Transformations
$\Bm{H}$ can be used.
These transformations are reflections on the plane
spanned by a vector $\Bv{v}_n \in \mathbb{R}^n$. To decompose a $D$-by-$Q$ matrix, $Q$ 
of such transformations are needed and the
resulting  \unitary matrix $\Bm{Q}$ can be written as 
\begin{equation}
 \Bm{Q} = \Bm{H}_D(\Bv{v}_D) \Bm{H}_{D-1}(\Bv{v}_{D-1}) ... \Bm{H}_{D-Q+1}(\Bv{v}_{D-Q+1}).
 \label{QbyHouseHolder}
\end{equation}
To construct $\Bm{H}_n$, we define $\tilde{\Bm{H_n}}(\Bv{v}_n)$ as
\begin{equation}
 \tilde{\Bm{H_n}}(\Bv{v}_n) = -\text{sgn}(\Bv{v}_{n1})\left( 
 \Bm{I} - 2 \Bv{u}_n\Bv{u}_n^T\right) 
 \in \mathbb{R}^{n \times n},
\end{equation}
where
\begin{equation}
  \label{Householder_un}
 \Bv{u}_n = \frac{\Bv{v}_n + \text{sgn}(\Bv{v}_{n1}) \|\Bv{v}_n\| \Bv{e}_1}
 {\lVert \Bv{v}_n + \text{sgn}(\Bv{v}_{n1}) \|\Bv{v}_n\| \Bv{e}_1 \rVert}
\end{equation}
and construct $\Bm{H}_n$ by
\begin{equation}
 \Bm{H}_n = 
 \begin{pmatrix}
  \Bm{I} & \Bm{0} \\
  \Bm{0} &  \tilde{\Bm{H_n}}
  \label{householder}
 \end{pmatrix}.
\end{equation}
The algorithm for the QR-decomposition constructs a suitable sequence
of vectors $\Bv{v}_D, \ldots, \Bv{v}_{D-Q+1}$ based on the entries of
the successively rotated columns of $\Bm{Z}$. The choice of the sign fulfills the requirements mentioned 
earlier, i.e. the diagonal elements of $\Bm{R}$ are positive.

Here, we are only interested in the resulting \unitary matrix
$\Bm{Q}$. In particular, by randomly drawing vectors $\Bm{v}$, we
obtain a random \unitary matrix via (\ref{QbyHouseHolder}).
The following theorem tells us which distribution the $\Bm{v}$s
need to have in order to get an \unitary matrix distributed according to 
the Haar measure \citep{mezzadri2007}.
\begin{theorem}
 Let $\Bv{v}_D, \Bv{v}_{D-1}, ..., \Bv{v}_1$ be uniformly distributed 
 on the unit spheres $\mathbb{S}^{D-1}, ..., \mathbb{S}^{0}$ respectively,
 where $\mathbb{S}^{n-1}$ is the unit sphere in $\mathbb{R}^n$. Furthermore,
 let $\Bm{H}_n(\Bv{v}_n)$ be the $n$-th Householder transformation
 as defined in equation (\ref{householder})
 The product
 \begin{equation}
  \Bm{Q} = \Bm{H}_D(\Bv{v}_D) \Bm{H}_{D-1}(\Bv{v}_{D-1}) ... \Bm{H}_1(\Bv{v}_1)
 \end{equation}
 is a random \unitary matrix with distribution given by the Haar measure
 on $O(D)$.
 \label{orthSphere}
\end{theorem}
A corresponding draw from the Stiefel manifold $\mathcal{V}_{Q,D}$ can be obtained by just
taking the first $Q$ columns of $\Bm{Q}$. Alternatively, this is achieved
by drawing vectors $\Bv{v}_D, \ldots, \Bv{v}_{D-Q+1}$ uniformly from the respective unit sphere
and construct the unitary matrix from the corresponding Householder transformations.
By the theorem above, this gives the Haar measure on $\mathcal{V}_{Q,D}$ as the Householder
transformations $\Bm{H}_{D-Q}, \ldots, \Bm{H}_{1}$ only effect the columns $D-Q, \ldots, 1$
by (\ref{householder}).

\section{Unique PPCA}
\label{UBPCA}

Here, we connect the results from the previous section to obtain a
model for PPCA without the latent space symmetry that plaques the
original formulation.  We start by sampling uniformly from the unit
spheres $\mathbb{S}^{n-1}$.  This is most easily achieved by drawing
an i.i.d. standard Gaussian vector of dimension $n$ and normalizing
its length. Note that this spends an additional parameter compared to
the dimensionality $n-1$ of the unit sphere $\mathbb{S}^{n-1}$.  Yet,
the $n$-dimensional standard Gaussian distribution is easy to sample
from and the vector length is sufficiently constrained by the Gaussian
prior such that the sampler can move around the unit sphere
effectively. The same parameterization is employed, for example, by Stan
in order to support a \verb|unit_vector| data type \citep{StanManual}.
Furthermore, the Householder transformation does not require the vectors $\Bv{v}$
to be of unit length as they are normalized anyways via (\ref{Householder_un}).
Thus, we construct the \unitary matrix $\Bm{U}$ distributed with the Haar measure
by successively transforming standard Gaussian random vectors.

Next, we sample the ordered singular values from the joint distribution
(\ref{pdf_sigma}). Again, this can be accomplished by transforming the
ordered vector $(\sigma_1, \ldots, \sigma_Q)$ to an unconstrained space
and correcting the joint density by the Jacobian determinant of the transformation.
We refer to \citet{StanManual} for details of this transformation. Finally,
from the \unitary matrix $\Bm{U}$ and the diagonal matrix $\Bm{\Sigma}$, 
we construct $\Bm{W}$ by $\Bm{W} = \Bm{U} \Bm{\Sigma}$, which 
we then use in the likelihood of the PPCA model. Overall, we obtain the following
generative model:
\begin{align}
  \Bv{v}_{D}, \ldots, \Bv{v}_{D-Q+1} &\sim \mathcal{N}(0,\Bm{I}) \nonumber \\
  \Bv{\sigma} &\sim p(\Bv{\sigma}) \propto \mbox{ eq. (\ref{pdf_sigma})} \nonumber \\
  \Bv{\mu} &\sim p(\Bv{\mu}) \mbox{, e.g. a broad Gaussian} \nonumber \\
  \Bm{U} &= \prod_{q=1}^Q \Bm{H}_{D-q+1}(\Bv{v}_{D-q+1}) \nonumber \\
  \Bm{\Sigma} &= \text{diag}(\Bv{\sigma}) \nonumber \\
  \Bm{W} &= \Bm{U} \Bm{\Sigma} \nonumber \\
  \sigma_{\text{noise}} &\sim p(\sigma_{\text{noise}}) \nonumber \\
  \Bm{Y} &\sim \prod_{n=1}^N \mathcal{N}\left( \Bm{Y}_{n,:} |
           \Bv{\mu}, \Bm{W}\Bm{W}^T +\sigma_{\text{noise}}^2 \Bm{I} \right). \nonumber
\end{align}

Note that this model defines the same distribution as the
corresponding model with a standard Gaussian prior
$\Bm{W} \sim \mathcal{N}(\Bv{0}, \Bm{I})$. In both cases, the sampling
distribution is governed by the Wishart distributed matrix
$\Bm{W} \Bm{W}^T$, even though the distribution on $\Bm{W}$ is actually
different. We implemented both models in the probabilistic programming
language Stan. The code for the simulations is available on Github:
\href{https://github.com/RSNirwan/HouseholderBPCA}
{https://github.com/RSNirwan/HouseholderBPCA}.


Compared to previous approaches based on parameterizing the Stiefel
manifold in terms of Givens rotations \citep{pourzanjani2017}, our
model has the following advantages: First, our Householder parameters $\Bv{v}$
 are unconstrained, in contrast to the
angular parameters of Givens rotations where the sampler might hit the
boundary of the space. Secondly, we avoid the computationally demanding
computation of the Jacobian determinant \citep{ron2014}. 
Compared to other approaches employing the matrix von Mises-Fisher
distribution \citep{Hoff2009,Smidl2007}, our model has the following advantages:
First, we do not require conditional conjugacy allowing non-linear
extensions of Bayesian PPCA as illustrated in section \ref{GPLVM}.
Secondly, we do not need to resort to rejection sampling or variational approximations.
Similar to these approaches,
our parameterization introduces a combinatorial symmetry by the sign
ambiguity of the SVD. This is akin to the label switching problem in
Gaussian mixture models which usually poses few problems with the sampler
simply getting stuck in a specific mode. Here, in order to compare
results across different modes, we postprocess each sample such that
the first entry of each column of $\Bv{U}$ is made positive. In the following,
we illustrate our model on some data sets and discuss possible extensions
to non-linear models where similar symmetries arise.

\subsection{Model Comparison}

\subsubsection{Synthetic Dataset}

Here, we build our own synthetic dataset with known parameters
and the goal is to reconstruct the parameter values.
For $(N, D, Q) = (150, 5, 2)$ we sample $\Bm{X} 
\in \mathbb{R}^{N \times D}$ from a standard 
normal distribution and construct $\Bm{W}$ by $\Bm{W}
= \Bm{U} \Bm{\Sigma} \in \mathbb{R}^{D \times Q}$,
where $\Bm{U}$ is sampled from the Stiefel manifold
with Haar measure and
we specify $\Bm{\Sigma} = \text{diag}(\sigma_1, \sigma_2)$,
where $(\sigma_1, \sigma_2) = (3.0, 1.0)$.
Then, we get the observation $\Bm{Y} = \Bm{X} \Bm{W}^T + 
\Bm{\epsilon}$, where $\Bm{\epsilon}$ denotes the 
noise sampled from a zero mean Gaussian
with a standard deviation of 0.01.

\begin{figure}
 \centering
 \includegraphics[width=0.49\textwidth]{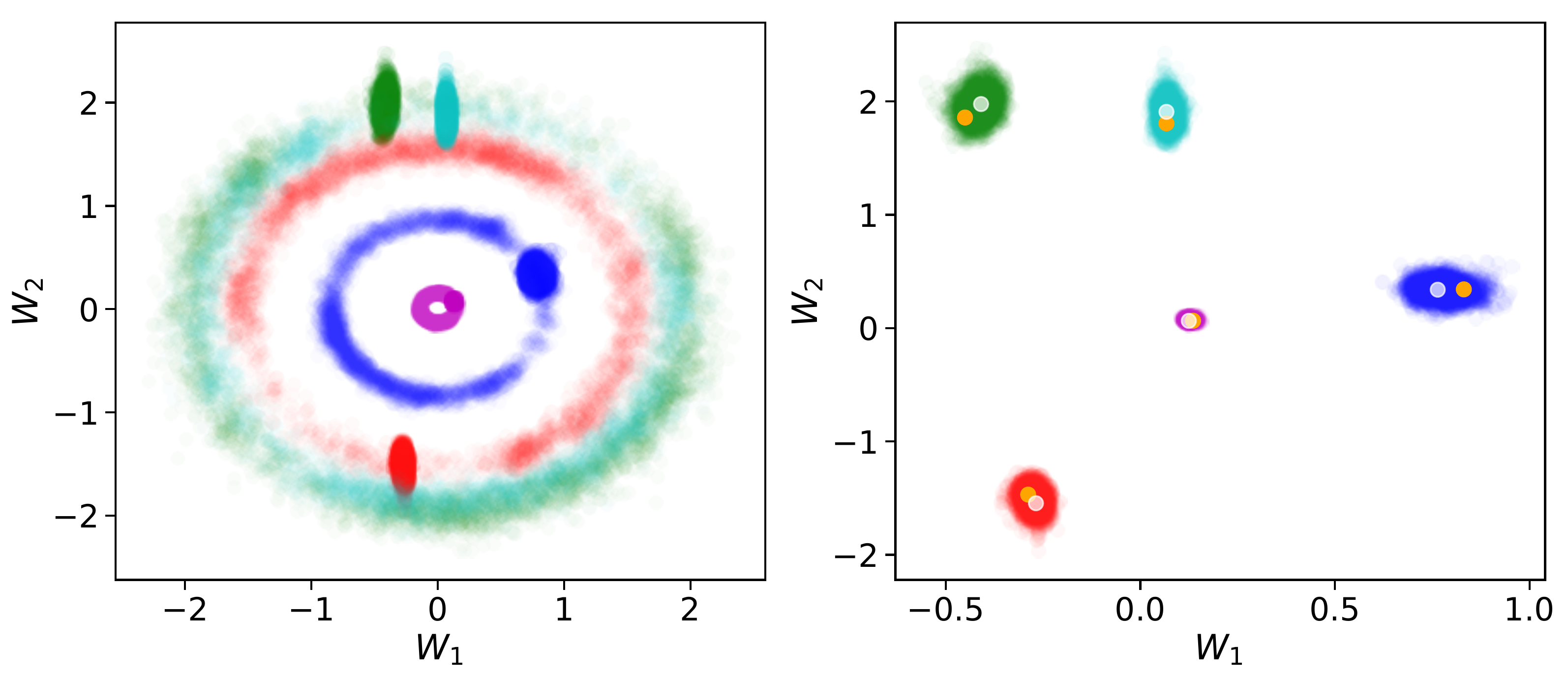}
 \vspace{-0.6cm}
 \caption{Results for synthetic dataset. 
 Left: In the background, we see samples from the posterior of
 the standard parameterization. Samples from the $\Bm{U}$ and $\Bm{\Sigma}$
 mapped to $\Bm{W} = \Bm{U}\Bm{\Sigma}$ are shown in a darker color. 
 Right: Comparison of the suggested model to
 classical PCA. White dots are the classical PCA solution and
 orange dots are the true values. 4000 samples for each row of $\Bm{W}$ are shown in 
 different colors.}
 \label{syntetic}
\end{figure}

The left plot in Figure \ref{syntetic} 
compares the Bayesian inference for the standard model, where 
we directly assume a standard Gaussian prior on $\Bm{W}$ with our
suggested model, where we parameterize $\Bm{W}$ by
$\Bm{U}$ and $\Bm{\Sigma}$ and infer the posterior 
distribution of both. 
As expected, the standard model has the
rotational symmetry in $\Bm{W}$, whereas our model takes out the
rotation\footnote{The additional sign ambiguity of the columns of $\Bm{U}$
is removed afterwards by post processing each sample.}. On the right side (Figure \ref{syntetic}),
we compare the posterior distribution of our model with the 
classical PCA solution and as expected with the low observation noise, the solutions are quite 
similar. In contrast to classical PCA, the Bayesian
approach provides a distribution for our parameters, accessing
estimation uncertainty of the solution as well.

\begin{figure}
 \centering
 \vspace{-0.05cm}
 \includegraphics[width=0.40\textwidth]{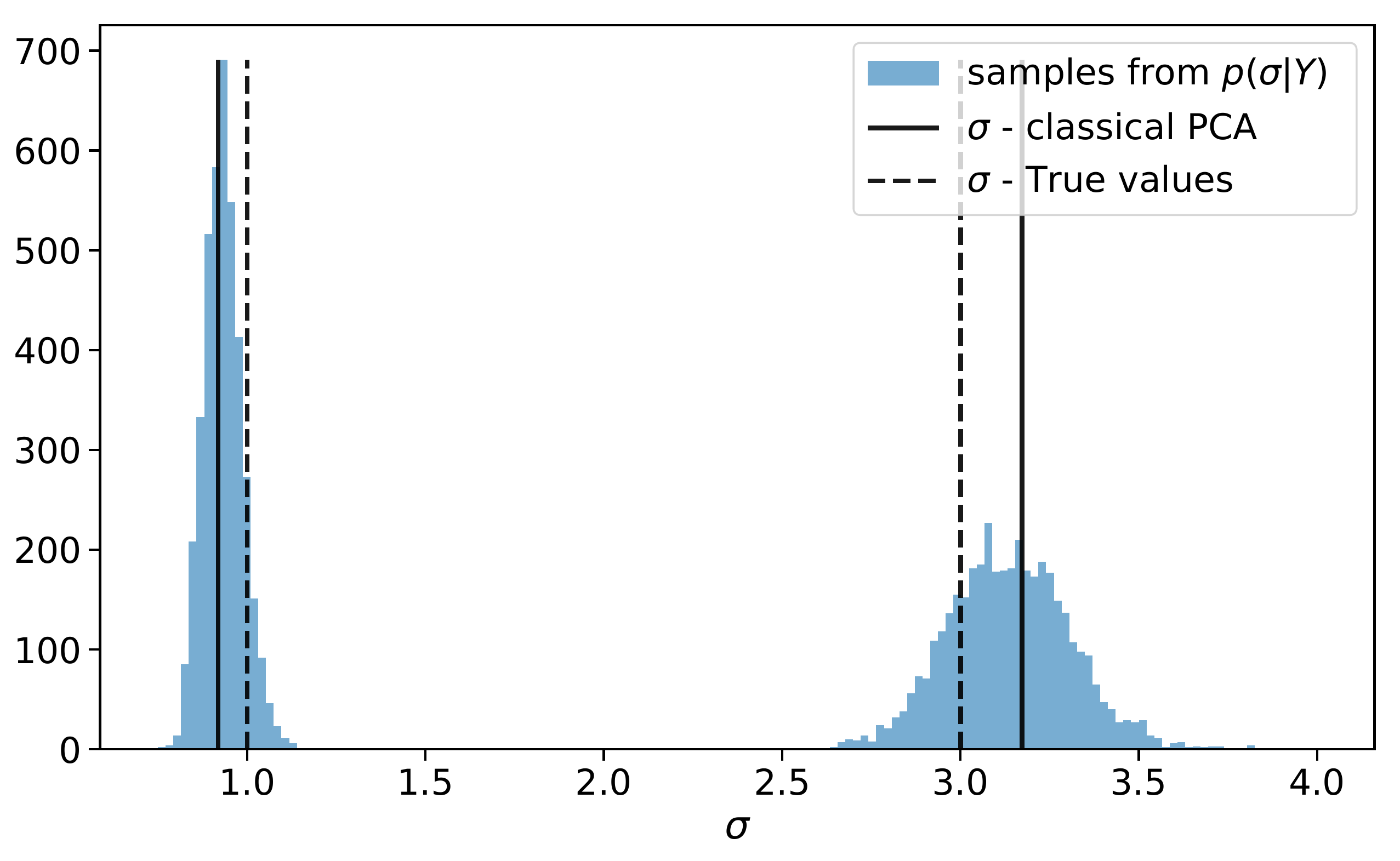}
 \vspace{-0.3cm}
 \caption{Histogram of the samples from the posterior distribution 
 of $\Bm{\Sigma} = \text{diag}(\sigma_1,
 \sigma_2)$. The vertical solid black lines are the classic PCA solution 
 and dashed lines are the true values $(3.0, 1.0)$,
 which generated the data.}
 \label{syntetic_sigma}
\end{figure}

Figure \ref{syntetic_sigma} shows the posterior distribution
of $\Bm{\Sigma}$, that we fixed to $\text{diag}(3.0, 1.0)$. As we can see,
the true values are recovered quite well.

\subsubsection{Breast Cancer Wisconsin Dataset}

\begin{figure}
 \centering
 \includegraphics[width=0.49\textwidth]{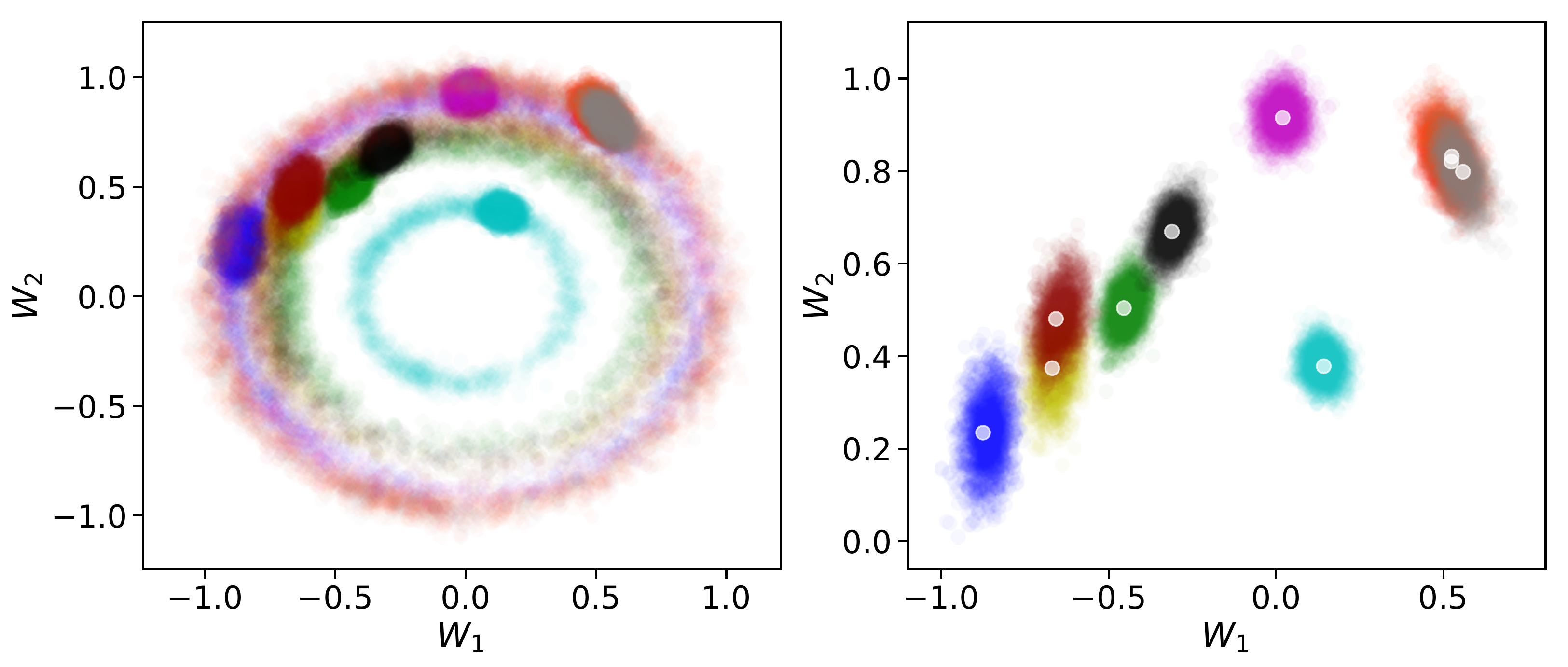}
 \vspace{-0.55cm}
 \caption{Results for the Breast Cancer Wisconsin dataset. Left: Posterior
 samples of the standard parameterization of $\Bm{W}$ (in
 the background) and posterior samples of $\Bm{W}$ parameterized
 by $\Bm{U}$ and $\Bm{\Sigma}$ (foreground). Right: Posterior
 samples of our parameterization and white dots are the 
 classical PCA solution. 4000 samples for each row of $\Bm{W}$ are shown in 
 different colors. For clarity, we only show 10 randomly chosen rows instead 
 of all 30.}
 \label{breast_cancer}
\end{figure}

We tested the model on the Breast Cancer Wisconsin dataset as well. The dataset
was downloaded from the Python toolbox scikit-learn \citep{scikitlearn2011}
and contains 569 labeled datapoints with 30 features. 
We neglect the labels and take the standardized $569 \times 30$ matrix as
the input to our model. For visualization purposes, we 
again map the data to $Q=2$. Figure \ref{breast_cancer} shows the
performance of different models. In the left plot, we
see the posterior samples for the direct parameterization of $\Bm{W}$
and the parameterization of $\Bm{W}$ by $\Bm{U}$ and $\Bm{\Sigma}$
and again, our model is able to uniquely identify $\Bm{W}$.
The right plot contains the uniquely identified $\Bm{W}$
and the classical PCA solution (white dots).
Again, our model enriches the classical PCA solution with uncertainty estimates.
Samples were drawn from 4 independent chains which all
converged to the same solution.

\subsection{Computational Complexity}

Compared to probabilistic PCA we compute the $D$x$Q$ loading matrix 
$\Bm{W}$ from parameters $\Bv{v}$ for the principal rotation and 
$\Bv{\sigma}^2$ for the principal variances. This is achieved via a 
sequence of $Q$ Householder transformations which, as detailed in 
\citet{ron2014}, can be achieved in $\mathcal{O}(DQ^2)$ steps. The 
same complexity applies when computing gradients with respect to the 
model parameters. Overall, this increases the total computation time 
of probabilistic PCA by a factor of $Q$, i.e. the number of latent 
dimensions, compared to the standard parameterization. 
The scaling with respect to the number of data points is unchanged.

When sampling from the model, the adaptive NUTS sampler, which is 
implemented in Stan, needs fewer leap-frog steps per sample because 
the rotational symmetry is removed from the posterior and does not 
need to be explored. Numerically, this leads to slightly lower wall 
clock times overall. From this perspective, our implementation 
is as efficient as other sampling methods proposed for 
probabilistic PCA or Bayesian factor analysis. 
If scalability becomes an issue, our model can easily be used 
together with more scalable approximation methods such 
as variational Bayes.

\citet{pourzanjani2017} proposed an alternative parameterization 
which removes the rotational symmetry by means of Givens rotations. 
This approach requires a Jacobian adjustment to account for the 
change of measure in order to ensure the uniform Haar measure. 
This requires the computation of the absolute determinant of an 
$D \cdot Q$ dimensional matrix. Thus, compared to their 
approach we achieve a substantial speedup as, according to 
theorem \ref{orthSphere},
no Jacobian adjustment is required in our model.

\subsection{Other than Gaussian Priors}

The prior in the paper has been chosen to aid comparison with standard 
(classical) PCA. Using the SVD, we decompose the prior into a rotation, i.e. the 
principal axis, and a diagonal matrix, containing the principal 
components/variances. Thus, the parameters of our model are easily 
and well interpretable. For the Gaussian prior, which is often chosen 
in Bayesian factor analysis to support Gibbs sampling, we show that 
the rotation and variances are independent. Furthermore, the rotation 
is uniformly distributed according to the Haar measure.
  
Other priors of this structure can be constructed without issues. 
In particular, the interpretation of the variance parameters 
$\Bv{\sigma}^2$ as the principal components helps to this end. 
For instance, automatic relevance determination is readily 
accomplished by putting a shrinkage/sparsity prior on the 
principle components. Interestingly, such a prior would be 
very different from priors arising from imposing sparsity 
on the factor loadings as suggested in sparse Bayesian factor 
analysis \citep{Bhattacharya2011}.
In this case, the induced distribution on the principal 
components is still sparse - at least numerically, as we are not 
aware of any theoretical results. More importantly though, the 
induced rotation is not distributed uniformly anymore. 
Thus, such a prior gives rise to an implicit a-priori preference 
for certain principal axes. Note that this does not break the 
rotational symmetry of the model which arises from the rotation 
of the latent space. Instead it imposes a preference for 
certain directions in data space which, as we believe, is 
actually undesirable in many cases. We view it as a strength of 
our model that interpretable priors can be choosen independently 
for the rotation and variances.

\section{Extension to Non-linear Models}
\label{GPLVM}

\subsection{Gaussian Process Latent Variable Model}

Since our method works well for the linear PPCA model, the
question arises, whether we could as well improve the 
posterior sampling of non-linear dimensionality reduction
techniques. \citet{Lawrence2005} introduced an extension of 
PPCA called the Gaussian Process Latent Variable Model
(GP-LVM). The model starts with
the dual formulation of PPCA. In this case, instead of marginalizing over
$\Bm{X}$ in equation (\ref{linear_eq_y}), we marginalize over
$\Bm{W} \sim \mathcal{N}(\Bv{0}, \Bm{I})$. The resulting likelihood then takes the form 

\begin{equation}
 p(\Bm{Y}|\Bm{X}) = \prod_{d=1}^{D} \mathcal{N}(\Bm{Y}_{:,d} | \Bv{\mu}, 
 \Bm{K} + \sigma^2 I),
 \label{gplvm_likelihood}
\end{equation}
where $\Bm{K} = \Bm{X}\Bm{X}^T$ and $\Bm{K}_{ij} = 
\Bm{X}_{i,:}^T\Bm{X}_{j,:}$.
Identifying this with the covariance kernel of Gaussian process \citep{Rasmussen2006},
the model is generalized by replacing the linear kernel $\Bm{K}_{ij}$
with a non-linear one. In this way, a non-linear 
dimensionality reduction method is obtained, the GP-LVM.
One of the most used covariance kernels is the exponentiated 
quadratic kernel \citep{Rasmussen2006}
\begin{equation}
 k_{\text{SE}}(\Bv{x}, \Bv{x}') = 
 \sigma_{\text{SE}}^2 \exp(-0.5 ||\Bv{x}- \Bv{x}'||^2_2/l^2), 
 \label{cov_function}
\end{equation}
where $\sigma_{\text{SE}}^2$ is the kernel variance and $l$ the
lengthscale.
We will use this covariance function in our analysis as well.

As suggested by 
\citet{Lawrence2005}, we can optimize equation (\ref{gplvm_likelihood})
$\log p(\Bm{Y}|\Bm{X})$ with respect to the latent positions
and the hyperparameters.
Optimization can easily lead to overfitting and a fully Bayesian
treatment of the model would be preferable. 
Therefore, we need a prior distribution $p(\Bm{X})$ 
on $\Bm{X}$ and using Bayes rule, the posterior is given as $p(\Bm{X}|\Bm{Y}) = 
p(\Bm{Y}|\Bm{X})p(\Bm{X})/p(\Bm{Y})$. As in the case of PPCA
the posterior is not analytically tractable and, as before, we resort to
sampling in order to approximate it. Note that the kernel $k_{SE}$ only
depends on the distance between points $\Bv{x}$ and $\Bv{x'}$ and thus
the model likelihood is again invariant under rotations of the latent space $\Bm{X}$.

\subsection{Model Comparison}

To test our model in the non-linear case, we again take the
Breast Cancer Wisconsin dataset. For the GP-LVM the 
input is the transposed matrix $\Bm{Y} \in \mathbb{R}^{N \times D}$,
where $N=30$ and $D=569$. We standardized the data and set 
$\sigma_{\text{SE}}$ and $l$ to one and only sample the latent 
space. First, we fit the
data with the standard parameterization, where we directly 
have the $\Bm{X}$ as parameters and then we reparameterize $\Bm{X}$ 
by $\Bm{U}$ and $\Bm{\Sigma}$ and sample $\Bv{v}$ and
$\Bv{\sigma}$ as described in sections \ref{first_part} and
\ref{second_part}.

\begin{figure}
 \centering
 \vspace{-0.1cm}
 \includegraphics[width=0.47\textwidth]{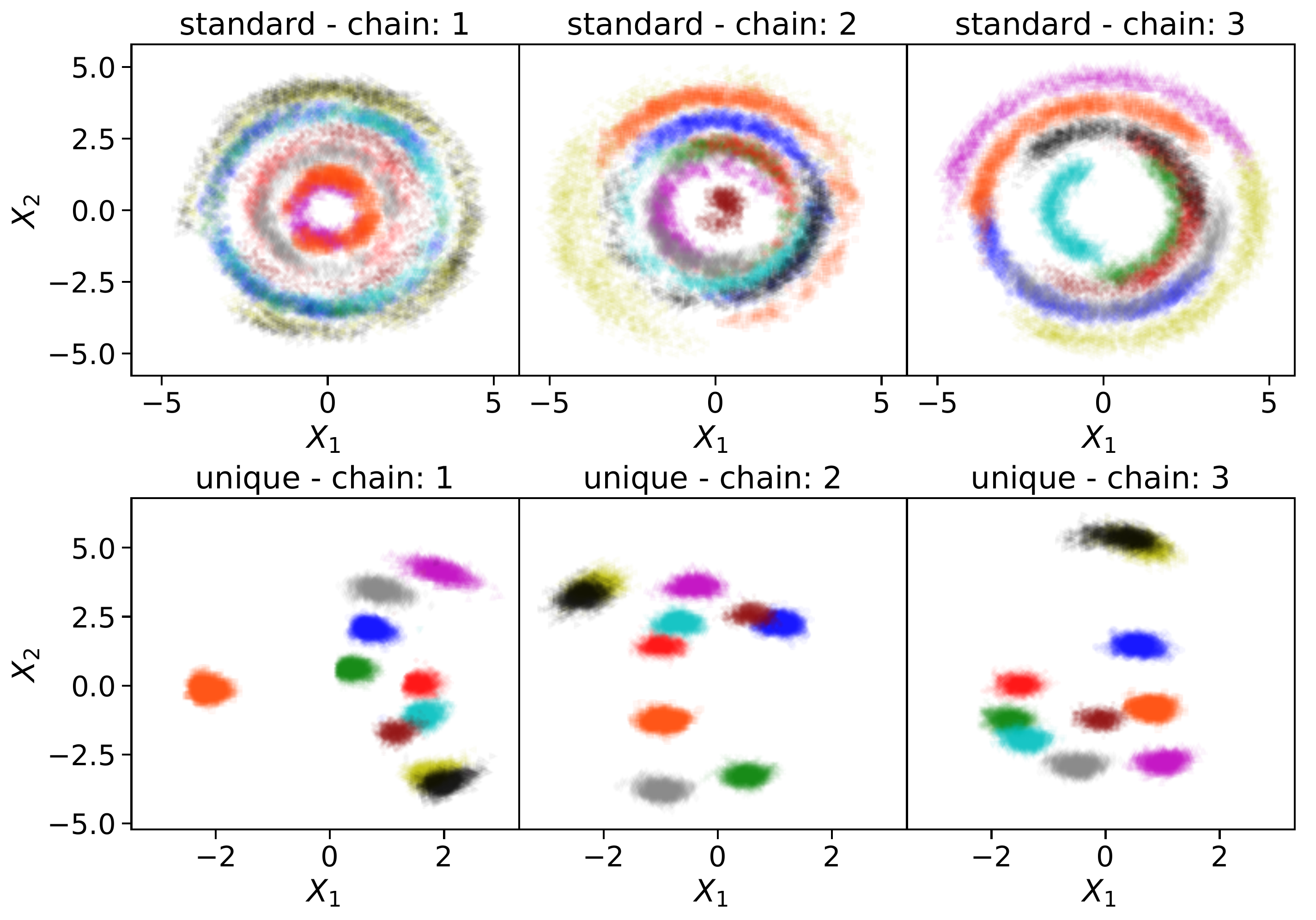}
 \vspace{-0.15cm}
 \caption{Posterior samples from for the GP-LVM. The top row shows
 the results for the standard parameterization for 
 $\Bm{X}$ and the bottom row shows the results of our 
 suggested parameterization. 2000 samples per chain for 
 each row of $\Bm{X}$ are shown in 
 different colors. Again, for clarity, we only show 10 
 randomly chosen rows instead of all 30.}
 \label{breast_cancer_kppca}
\end{figure}

Figure \ref{breast_cancer_kppca} shows the results for the 
exponentiated quadratic kernel for $Q=2$. 
As before, we used Stan to sample from the
posterior on three independent chains. The top row shows the 
samples from the different chains for the standard 
parameterization. One can clearly see the
rotational symmetry. The bottom row shows the samples 
with our suggested parameterization.  
As expected, there is no rotational symmetry anymore. 
However, due to the models complexity/flexibility many
local minima arise and the chains with both (the standard and 
our suggested) parameterizations do not converge to the same 
posterior.

\section{Conclusion}
\label{conclusion}

We suggested a new parameterization for the mapping $\Bm{W}$
in PPCA taking latent to observed points, which 
uniquely identifies the principal components even though
the likelihood and the posterior (for the Bayesian case)
are rotationally symmetric. We have shown how to parameterize the model
via the singular vectors and values of $\Bm{W}$ and how to set the prior
on the new parameters such that the model is not changed compared
to a standard Gaussian prior on $\Bm{W}$ directly. Furthermore,
we provided an efficient implementation via Householder transformations.

Thereafter, we tested the model on a synthetic and the Breast Cancer Wisconsin
dataset. Our model was able to uniquely reconstruct the true parameters
that created the synthetic data. On both datasets
our model provided similar results as classical PCA,
additionally containing the uncertainty of the estimates as well.
In the end, we showed that our method can also be used to remove the symmetry
in the latent space of a non-linear GP-LVM model. Overall, we believe that 
our approach, thanks to its known prior distribution and computational efficiency,
can be quite useful for latent space models with rotation symmetric likelihoods.

\section*{Acknowledgements}

The authors thank Dr. h.c. Maucher for funding their positions.


\bibliography{icml2019}
\bibliographystyle{icml2019}

\end{document}